\documentclass[11pt,a4paper]{article}
\usepackage[noend]{algpseudocode}
\usepackage{algorithm}
\usepackage{CJKutf8}
\usepackage[hyperref]{acl2021}
\usepackage{times}
\usepackage{latexsym}
\usepackage{booktabs}
\usepackage{multirow}
\usepackage{graphicx}
\usepackage{rotating}

\usepackage{subfigure}

\usepackage{tikz}
\usetikzlibrary{shapes.geometric}
\usepackage{pgfplots}
\pgfplotsset{compat=newest}
\pgfdeclareplotmark{mystar}{
\node[star,star point ratio=2.25,minimum size=6pt,
      inner sep=0pt,draw=black,solid,fill=red] {};
}
\definecolor{royalblue}{RGB}{65,105,225}
\usepackage{amsmath}
\usepackage{graphicx}
\usepackage{float}
\usepackage{subfigure}
\usepackage{booktabs}
\usepackage{amssymb}
\usepackage{threeparttable}
\usepackage[disable]{todonotes}
\usepackage{pgfplots}
\usepackage{PGFPlotsTable}
\usepackage{scalefnt}

\usepackage{microtype}

\aclfinalcopy %
\def\aclpaperid{2269} %

\title{Accelerating BERT Inference for Sequence Labeling via Early-Exit}

\author{
Xiaonan Li, Yunfan Shao, Tianxiang Sun, Hang Yan, Xipeng Qiu\thanks{\ \  Corresponding author.}, Xuanjing Huang \\
Shanghai Key Laboratory of Intelligent Information Processing, Fudan University \\
School of Computer Science, Fudan University \\
\{lixn20, yfshao19, txsun19, hyan19, xpqiu, xjhuang\}@fudan.edu.cn
}

\date{}

\begin{document}
\maketitle
\begin{CJK}{UTF8}{gbsn}
\begin{abstract}

Both performance and efficiency are crucial factors for sequence labeling tasks in many real-world scenarios. Although the pre-trained models (PTMs) have significantly improved the performance of various sequence labeling tasks, their computational cost is expensive. To alleviate this problem, we extend the recent successful early-exit mechanism to accelerate the inference of PTMs for sequence labeling tasks.
However, existing early-exit mechanisms are specifically designed for sequence-level tasks, rather than sequence labeling. In this paper, we first propose \textbf{SENTEE}: a simple extension of \textbf{SENT}ence-level \textbf{E}arly-\textbf{E}xit for sequence labeling tasks. To further reduce computational cost, we also propose \textbf{TOKEE}: a \textbf{TOK}en-level \textbf{E}arly-\textbf{E}xit mechanism that allows partial tokens to exit early at different layers. Considering the local dependency inherent in sequence labeling, we employed a window-based criterion to decide for a token whether or not to exit. The token-level early-exit brings the gap between training and inference, so we introduce an extra self-sampling fine-tuning stage to alleviate it.
The extensive experiments on three popular sequence labeling tasks show that our approach can save up to 66\%$\sim$75\% inference cost with minimal performance degradation. Compared with competitive compressed models such as DistilBERT, our approach can achieve better performance under the same speed-up ratios of 2$\times$, 3$\times$, and 4$\times$.\footnote{Our implementation is publicly available at \href{https://github.com/LeeSureman/Sequence-Labeling-Early-Exit}{https:// github.com/LeeSureman/Sequence-Labeling-Early-Exit}.}

\end{abstract}

\section{Introduction}

Sequence labeling plays an important role in natural language processing (NLP). Many NLP tasks can be converted to sequence labeling tasks, such as named entity recognition, part-of-speech tagging, Chinese word segmentation and Semantic Role Labeling.
These tasks are usually fundamental and highly time-demanding, therefore, apart from performance, their inference efficiency is also very important.

The past few years have witnessed the prevailing of pre-trained models (PTMs)~\citep{DBLP:journals/corr/abs-2003-08271} on various sequence labeling tasks~\citep{nguyen-etal-2020-bertweet,ke2020unified,tian-etal-2020-joint-chinese,mengge-etal-2020-coarse1}. Despite their significant improvements on sequence labeling, they are notorious for enormous computational cost and slow inference speed, which hinders their utility in real-time scenarios or mobile-device scenarios.

\begin{figure}[!t]
  \centering

  \subfigure[Early-Exit for Text Classification]{
    \includegraphics[width=0.45\textwidth]{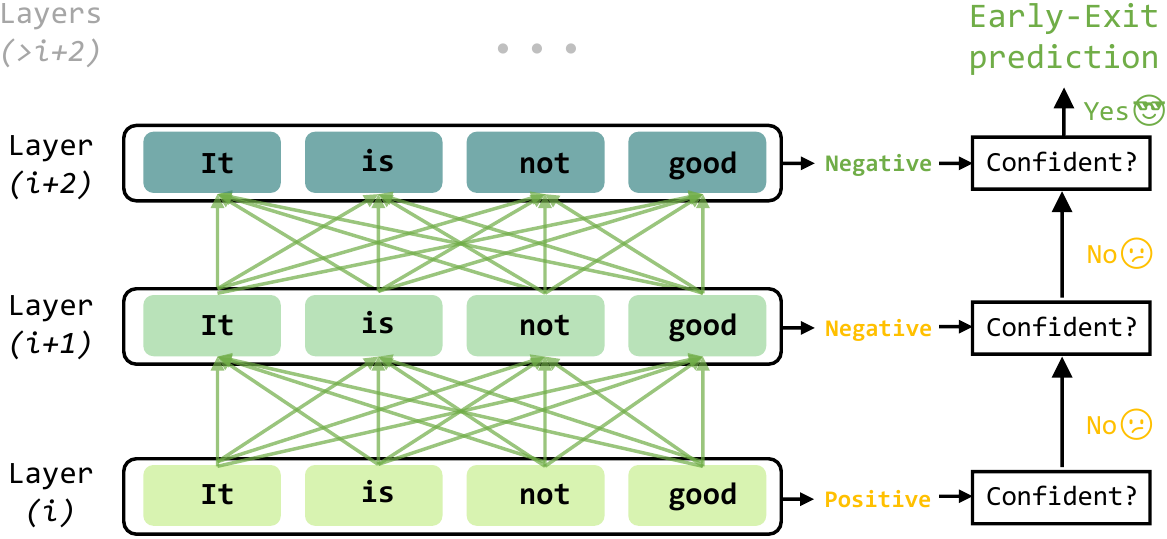}
    \label{text_cls_ee_fig}
  }
  \subfigure[Sentence-Level Early-Exit for Sequence Labeling]{
    \includegraphics[width=0.45\textwidth]{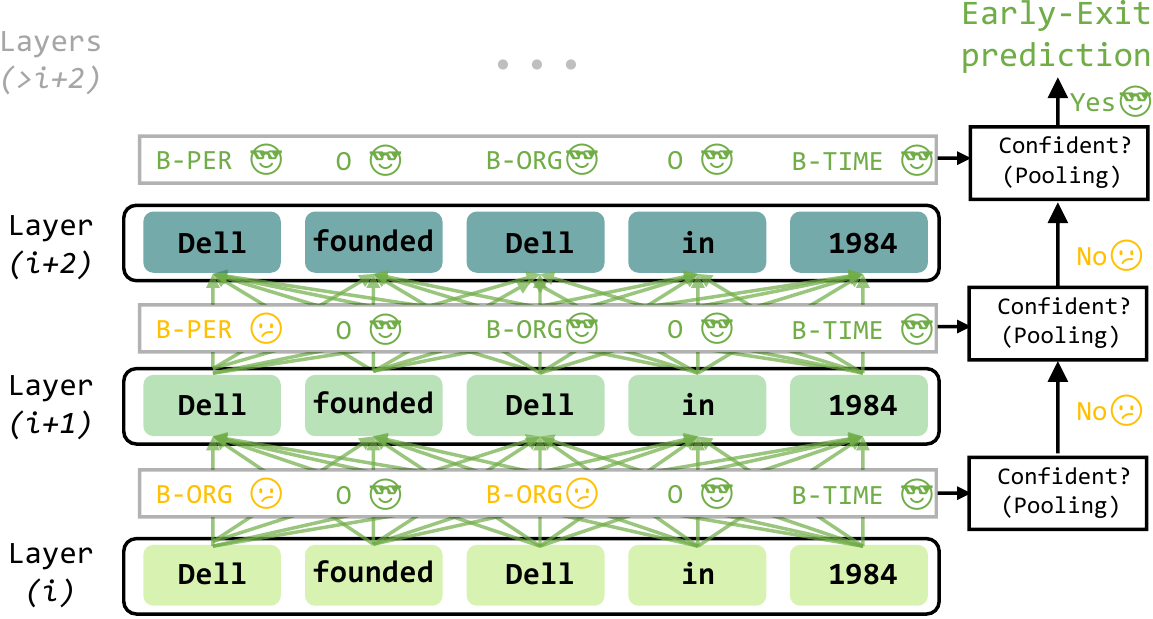}
    \label{sentence_ee_fig}
  }
  \subfigure[Token-Level Early-Exit for Sequence Labeling]{
    \includegraphics[width=0.45\textwidth]{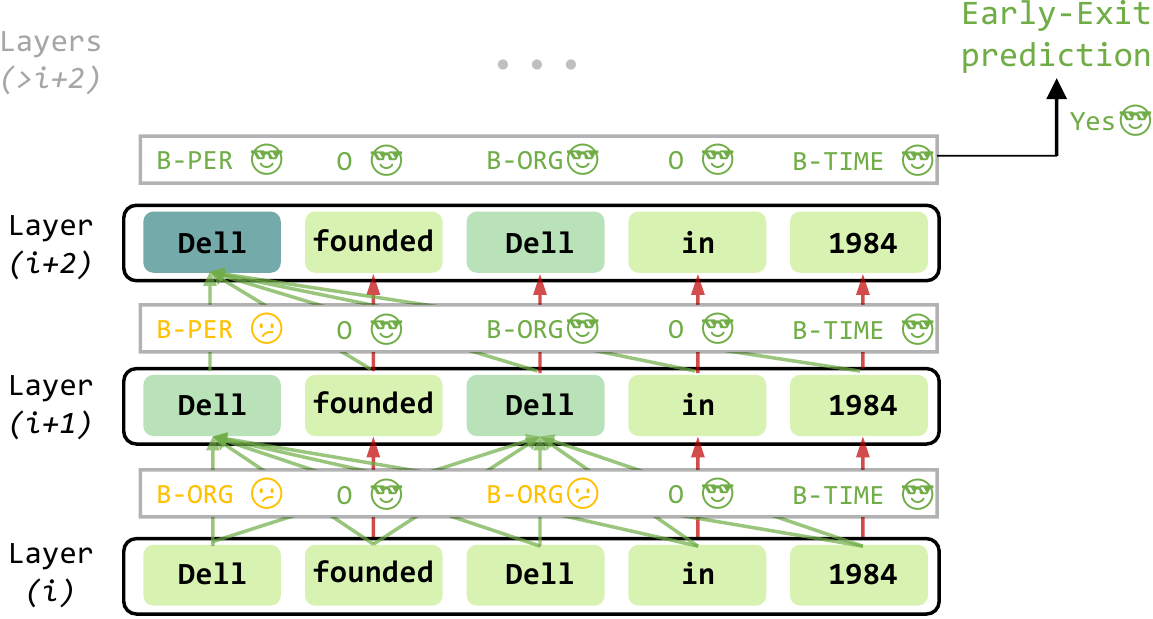}
    \label{token_ee_fig}
  }
  \caption{Early-Exit for Text Classification and Sequence Labeling, where \includegraphics[width=.35cm]{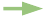} represents the self-attention connection, \includegraphics[width=.35cm]{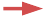} represents simply copying, \includegraphics[width=.35cm]{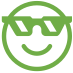} and \includegraphics[width=.35cm]{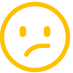} represents the confident and uncertain prediction. The darker the hidden state block, the deeper layer it is from. Due to space limit, the window-based uncertainty is not reflected.}
\end{figure}

Recently, early-exit mechanism~\citep{liu-etal-2020-fastbert,xin-etal-2020-deebert,schwartz-etal-2020-right,zhou-bert-patience} has been introduced to accelerate inference for large-scale PTMs. In their methods, each layer of the PTM is coupled with a classifier to predict the label for a given instance. At inference stage, if the prediction is confident \footnote{In this paper, confident prediction indicates that the uncertainty of it is low.} enough at an earlier time, it is allowed to exit without passing through the entire model. Figure~\ref{text_cls_ee_fig} gives an illustration of early-exit mechanism for text classification.
However, most existing early-exit methods are targeted at sequence-level prediction, such as text classification, in which the prediction and its confidence score are calculated over a sequence. Therefore, these methods cannot be directly applied to sequence labeling tasks, where the prediction is token-level and the confidence score is required for each token.

In this paper, we aim to extend the early-exit mechanism to sequence labeling tasks. First, we proposed the \textit{SENTence-level Early-Exit} (SENTEE), which is a simple extension of existing early-exit methods. SENTEE allows a sequence of tokens to exit together once the maximum uncertainty of the tokens is below a threshold. Despite its effectiveness, we find it redundant for most tokens to update the representation at each layer. Thus, we proposed a \textit{TOKen-level Early-Exit} (TOKEE) that allows part of tokens that get confident predictions to exit earlier. Figure~\ref{sentence_ee_fig} and \ref{token_ee_fig} illustrate our proposed SENTEE and TOKEE. Considering the local dependency inherent in sequence labeling tasks, we decide whether a token could exit based on the uncertainty of a window of its context instead of itself. For tokens that are already exited, we do not update their representation but just copy it to the upper layers. However, this will introduce a train-inference discrepancy. To tackle this problem, we introduce an additional fine-tuning stage that samples the token's halting layer based on its uncertainty and copies its representation to upper layers during training. We conduct extensive experiments on three sequence labeling tasks: NER, POS tagging, and CWS. Experimental results show that our approach can save up to 66\% $\sim$75\% inference cost with minimal performance degradation. Compared with competitive compressed models such as DistilBERT, our approach can achieve better performance under speed-up ratio of 2$\times$, 3$\times$, and 4$\times$.

\section{BERT for Sequence Labeling}
Recently, PTMs~\citep{DBLP:journals/corr/abs-2003-08271} have become the mainstream backbone model for various sequence labeling tasks.
The typical framework consists of a backbone encoder and a task-specific decoder.

\paragraph{Encoder}
In this paper, we use BERT~\citep{devlin-etal-2019-bert} as our backbone encoder
.
The architecture of BERT consists of multiple stacked Transformer layers~\citep{att-is-need}.

Given a sequence of tokens $x_1,\cdots, x_N$,
the hidden state of $l$-th transformer layer is denoted by $\mathbf{H}^{(l)} = [\mathbf{h}_1^{(l)}, \cdots, \mathbf{h}_N^{(l)}]$, and $\mathbf{H}^{(0)}$ is the BERT input embedding.

\paragraph{Decoder}

Usually, we can predict the label for each token according to the hidden state of the top layer.
The probability of labels is predicted by
\begin{align}
    \mathbf{P} &= f(\mathbf{W}\mathbf{H}^{(L)}) \in \mathbb{R}^{N\times C},
    \label{cls_prob_eq}
\end{align}
where $N$ is the sequence length, $C$ is the number of labels, $L$ is the number of BERT layers, $\mathbf{W}$ is a learnable matrix, and $f(\cdot)$ is a simple softmax classifier or conditional random field (CRF)~\citep{Lafferty:2001:CRF:645530.655813}. Since we focus on inference acceleration and PTM performs well enough on sequence labeling without CRF \citep{devlin-etal-2019-bert}, we do not consider using such a recurrent structure.

\section{Early-Exit for Sequence Labeling}

The inference speed and computational costs of PTMs are crucial bottlenecks to hinder their application in many real-world scenarios. In many tasks, the representations at an earlier layer of PTMs are usually adequate to make a correct prediction. Therefore, early-exit mechanisms~\citep{liu-etal-2020-fastbert,xin-etal-2020-deebert,schwartz-etal-2020-right,zhou-bert-patience} are proposed to dynamically stop inference on the backbone model and make prediction with intermediate representation.

However, these existing early-exit mechanisms are built on sentence-level prediction and unsuitable for token-level prediction in sequence labeling tasks.
In this section, we propose two early-exist mechanisms to accelerate the inference for sequence labeling tasks.

\subsection{Token-Level Off-Ramps}

To extend early-exit to sequence labeling, we couple each layer of the PTM with token-level s that can be simply implemented as a linear classifier. Once the off-ramps are trained with the golden labels, the instance has a chance to be predicted and exit at an earlier time instead of passing through the entire model.

Given a sequence of tokens $X=x_1,\cdots,x_N$, we can make predictions by the injected off-ramps at each layer. For an off-ramp at $l$-th layer, the label distribution of all tokens is predicted by
\begin{align}
     \mathbf{P}^{(l)} &= f^{(l)}(X;\theta) \\
    &= \operatorname{softmax}(\mathbf{W}\mathbf{H}^{(l)}),
    \label{cls_prob_eq}
\end{align}
where $\mathbf{W}$ is a learnable matrix, $f^{(l)}$ is the token-level off-ramp at $l$-th layer, $\mathbf{P}^{(l)}=[\mathbf{p}_1^{(l)}, \cdots, \mathbf{p}_N^{(l)}]$, $\mathbf{p}_n^{(l)} \in \mathbb{R}^{C}$, indicates the predicted label distribution at the $l$-th off-ramp for each token.

 \paragraph{Uncertainty of the Off-Ramp}
 With the prediction for each token at hand, we can calculate the \textit{uncertainty} for each token as follows,
\begin{equation}
    u_n^{(l)}=\frac{-\mathbf{p}^{(l)}_n \cdot \log\mathbf{p}^{(l)}_n}{\log \mathrm{C}},
\end{equation}
where $\mathbf{p}^{(l)}_n$ is the label probability distribution for the $n$-th token.

\subsection{Early-Exit Strategies}
In the following sections, we will introduce two early-exit mechanisms for sequence labeling, at sentence-level and token-level.

\subsubsection{SENTEE: Sentence-Level Early-Exit}
Sentence-Level Early-Exit (SENTEE) is a simple extension for sequential labeling tasks based on existing early-exit approaches. SENTEE allows a sequence of tokens to exit together if their uncertainty is low enough. Therefore, SENTEE is to aggregate the uncertainty for each token to obtain an overall uncertainty for the whole sequence. Here we perform a straight-forward but effective method, i.e., conduct max-pooling\footnote{We also tried average-pooling, but it brings drastic performance drop. We find that the average uncertainty over the sequence is often overwhelmed by lots of easy tokens and this causes many wrong exits of difficult tokens.}
over uncertainties of all the tokens,
\begin{equation}
    u^{(l)} = \operatorname{max}\{u^{(l)}_1, \cdots, u^{(l)}_n\},
\end{equation}
where $u^{(l)}$ represents the uncertainty for the whole sentence. If $u^{(l)} < \delta$ where $\delta$ is a pre-defined threshold, we let the sentence exit at layer $l$. The intuition is that only when the model is confident of its prediction for the most difficult token, the whole sequence could exit.

\subsubsection{TOKEE: Token-Level Early-Exit}
Despite the effectiveness of SENTEE (see Table~\ref{main_exp_table}), we find it redundant for most simple tokens to be fed into the deep layers. The simple tokens that have been correctly predicted in the shallow layer can not exit (under SENTEE) because the uncertainty of a small number of difficult tokens is still above the threshold. Thus, to further accelerate the inference for sequence labeling tasks, we propose a token-level early-exit (TOKEE) method that allows simple tokens with confident predictions to exit early.

\paragraph{Window-Based Uncertainty}
Note that a prevalent problem in sequence labeling tasks is the local dependency (or label dependency). That is, the label of a token heavily depends on the tokens around it. To that end, the calculation of the uncertainty for a given token should not only be based on itself but also its context. Motivated by this, we proposed a window-based uncertainty criterion to decide for a token whether or not to exit at the current layer. In particular, the uncertainty for the token $x_n$ at $l$-th layer is defined as
\begin{equation}
    u_n'^{(l)} = \operatorname{max}\{u^{(l)}_{n-k}, \cdots, u^{(l)}_{n+k}\},
\end{equation}
where $k$ is a pre-defined window size. Then we use $u_n'^{(l)}$ to decide whether the $n$ th token can exit at layer $l$, instead of $u_n^{(l)}$. Note that window-based uncertainty is equivalent to sentence-level uncertainty when $k$ equals to the sentence length.

\paragraph{Halt-and-Copy}

For tokens that have exited, their representation would not be updated in the upper layers, i.e., the hidden states of exited tokens are directly \textit{copied} to the upper layers.\footnote{For English sequence labeling, we use the first-pooling to get the representation of the word. If a word exits, we will halt-and-copy its all wordpieces.} Such a \textit{halt-and-copy} mechanism is rather intuitive in two-fold:
\begin{itemize}
\setlength{\itemsep}{1pt}%
\setlength{\parskip}{1pt}%
    \item \textbf{Halt.}~~If the uncertainty of a token is very small, there are also few chances that its prediction will be changed in the following layers. So it is redundant to keep updating its representation.
    \item \textbf{Copy.}~~If the representation of a token can be classified into a label with a high degree of confidence, then its representation already contains the label information. So we can directly copy its representation into the upper layers to help predict the labels of other tokens.
\end{itemize}

These exited tokens will not attend to other tokens at upper layers but can still be attended by other tokens thus part of the layer-specific query projections in upper layers can be omitted. By this, the computational complexity in self-attention is reduced from $\mathcal{O}(N^2d)$ to $\mathcal{O}(NMd)$, where $M\ll N$ is the number of tokens that have not exited. Besides, the computational complexity of the point-wise FFN can also be reduced from $\mathcal{O}(Nd^2)$ to $\mathcal{O}(Md^2)$.

The halt-and-copy mechanism is also similar to multi-pass sequence labeling paradigm, in which the tokens are labeled their in order of difficulty (easiest first).
However, the copy mechanism results in a train-inference discrepancy. That is, a layer never processed the representation from its non-adjacent previous layers during training. To alleviate the discrepancy, we further proposed an additional fine-tuning stage, which will be discussed in Section \ref{sec:ft-tokenEE}.

\subsection{Model Training}
In this section, we describe the training process of our proposed early-exit mechanisms.
\subsubsection{Fine-Tuning for SENTEE}
For sentence-level early-exit, we follow prior early-exit work for text classification to jointly train the added off-ramps. For each off-ramp, the loss function is as follows,
\begin{equation}
    \mathcal{L}_l= \sum_{n=1}^N
    H\left(y_n, f^{(l)}(X; \theta)_n\right),
\end{equation}
where $H$ is the cross-entropy loss function, $N$ is the sequence length. The total loss function for each sample is a weighted sum of the losses for all the off-ramps,
\begin{equation}
    \mathcal{L}_{\text{total}} = \frac{\sum _{l=1}^L w_l \mathcal{L}_l}{\sum_{l=1}^L w_l},
\end{equation}
where $w_l$ is the weight for the $l$-th off-ramp and $L$ is the number of backbone layers. Following \citep{zhou-bert-patience}, we simply set $w_l=l$. In this way, The deeper an off-ramp is, the weight of its loss is bigger, thus each off-ramp can be trained jointly in a relatively balanced way.

\subsubsection{Fine-Tuning for TOKEE}
\label{sec:ft-tokenEE}

Since we equip halt-and-copy in TOKEE, the common joint training off-ramps are not enough. Because the model never conducts halt-and-copy in training but does in inference. In this stage, we aim to train the model to use the hidden state from different previous layers but not only the previous adjacent layer, just like in inference.

\paragraph{Random Sampling}
A direct way is to uniformly sample halting layers of tokens. However, halting layers at the inference are not random but depends on the difficulty of each token in the sequence. So random sampling halting layers also causes the gap between training and inference.

\paragraph{Self-Sampling}
Instead, we use the fine-tuned model itself to sample the halting layers. For every sample in each training epoch, we will randomly sample a window size and threshold for it, and then we can conduct TOKEE on the trained model, under the window size and threshold, without halt-and-copy. Thus we get the exiting layer of each token, and we use it to re-forward the sample, by halting and copying each token in the corresponding layer. In this way, the exiting layer of a token can correspond to its difficulty. The deeper a token's exiting layer is, the more difficult it is. Because we sample the exiting layer using the model itself, we think the gap between training and inference can be further shrunk. To avoid over-fitting during further training, we prevent the training loss from further reducing, similar with the flooding mechanism used by \citet{pmlr-v119-ishida20a}. We also employ the sandwich rule to stabilize this training stage \cite{DBLP:conf/iccv/YuH19}. We compare self-sampling with random sampling in Section \ref{self-sampling-section}.

\begin{table*}[htbp]\scriptsize
  \setlength{\tabcolsep}{2.6mm}
  \centering
\begin{tabular}{@{}lccccccccccc@{}}
\toprule
 & \textbf{Task} & \multicolumn{10}{c}{\textbf{NER}} \\ \midrule[1pt]
 & \multicolumn{1}{c|}{\multirow{2}{*}{\textbf{\begin{tabular}[c]{@{}c@{}}Dataset/\\ Model\end{tabular}}}} & \multicolumn{2}{c|}{\textbf{CoNLL2003}} & \multicolumn{2}{c|}{\textbf{Twitter}} & \multicolumn{2}{c|}{\textbf{Ontonotes 4.0}} & \multicolumn{2}{c|}{\textbf{Weibo}} & \multicolumn{2}{c}{\textbf{CLUE NER}} \\
 & \multicolumn{1}{c|}{} & F1($\Delta$) & \multicolumn{1}{c|}{Speedup} & F1($\Delta$) & \multicolumn{1}{c|}{Speedup} & F1($\Delta$) & \multicolumn{1}{c|}{Speedup} & F1($\Delta$) & \multicolumn{1}{c|}{Speedup} & F1($\Delta$) & Speedup \\ \midrule
 & \multicolumn{1}{c|}{BERT} & 91.20 & \multicolumn{1}{c|}{1.00$\times$} & 77.97 & \multicolumn{1}{c|}{1.00$\times$} & 79.18 & \multicolumn{1}{c|}{1.00$\times$} & 66.15 & \multicolumn{1}{c|}{1.00$\times$} & 79.11 & 1.00$\times$ \\ \midrule
 & \multicolumn{1}{c|}{LSTM-CRF} & -0.29 & \multicolumn{1}{c|}{-} & -12.65 & \multicolumn{1}{c|}{-} & -7.37 & \multicolumn{1}{c|}{-} & -9.40 & \multicolumn{1}{c|}{-} & -7.75 & - \\ \midrule
\multicolumn{1}{c}{\parbox[t]{0mm}{\multirow{3}{*}{\rotatebox[origin=c]{90}{{$\sim$2X}}}}} & \multicolumn{1}{c|}{DistilBERT-6L} & -1.42 & \multicolumn{1}{c|}{2.00$\times$} & -1.79 & \multicolumn{1}{c|}{2.00$\times$} & -3.76 & \multicolumn{1}{c|}{2.00$\times$} & -1.95 & \multicolumn{1}{c|}{2.00$\times$} & -2.05 & 2.00$\times$ \\
 & \multicolumn{1}{c|}{SENTEE} & \textbf{+0.01} & \multicolumn{1}{c|}{2.07$\times$} & -0.82 & \multicolumn{1}{c|}{2.03$\times$} & -1.70 & \multicolumn{1}{c|}{1.94$\times$} & -0.23 & \multicolumn{1}{c|}{2.11$\times$} & \textbf{-0.22} & 2.04$\times$ \\
 & \multicolumn{1}{c|}{TOKEE} & -0.09 & \multicolumn{1}{c|}{2.02$\times$} & \textbf{-0.20} & \multicolumn{1}{c|}{2.00$\times$} & \textbf{-0.20} & \multicolumn{1}{c|}{2.00$\times$} & \textbf{-0.15} & \multicolumn{1}{c|}{2.10$\times$} & -0.24 & 1.99$\times$ \\
\midrule
\multicolumn{1}{c}{\parbox[t]{0mm}{\multirow{3}{*}{\rotatebox[origin=c]{90}{{$\sim$3X}}}}} & \multicolumn{1}{c|}{DistilBERT-4L} & -1.57 & \multicolumn{1}{c|}{3.00$\times$} & -3.37 & \multicolumn{1}{c|}{3.00$\times$} & -8.79 & \multicolumn{1}{c|}{3.00$\times$} & -5.92 & \multicolumn{1}{c|}{3.00$\times$} & -5.67 & 3.00$\times$ \\
 & \multicolumn{1}{c|}{SENTEE} & -0.75 & \multicolumn{1}{c|}{3.02$\times$} & -5.54 & \multicolumn{1}{c|}{3.03$\times$} & -8.90 & \multicolumn{1}{c|}{2.95$\times$} & -2.51 & \multicolumn{1}{c|}{2.87$\times$} & -6.17 & 2.98$\times$ \\
 & \multicolumn{1}{c|}{TOKEE} & \textbf{-0.32} & \multicolumn{1}{c|}{3.03$\times$} & \textbf{-1.32} & \multicolumn{1}{c|}{2.98$\times$} & \textbf{-1.48} & \multicolumn{1}{c|}{3.02$\times$} & \textbf{-0.70} & \multicolumn{1}{c|}{3.04$\times$} & \textbf{-0.92} & 2.91$\times$ \\ \midrule
\multicolumn{1}{c}{\parbox[t]{0mm}{\multirow{3}{*}{\rotatebox[origin=c]{90}{{$\sim$4X}}}}} & \multicolumn{1}{c|}{DistilBERT-3L} & -2.78 & \multicolumn{1}{c|}{4.00$\times$} & -5.08 & \multicolumn{1}{c|}{4.00$\times$} & -12.77 & \multicolumn{1}{c|}{4.00$\times$} & -9.05 & \multicolumn{1}{c|}{4.00$\times$} & -8.22 & 4.00$\times$ \\
 & \multicolumn{1}{c|}{SENTEE} & -6.51 & \multicolumn{1}{c|}{4.00$\times$} & -11.50 & \multicolumn{1}{c|}{4.14$\times$} & -14.40 & \multicolumn{1}{c|}{3.78$\times$} & -8.31 & \multicolumn{1}{c|}{3.86$\times$} & -14.95 & 3.91$\times$ \\
 & \multicolumn{1}{c|}{TOKEE} & \textbf{-1.44} & \multicolumn{1}{c|}{3.94$\times$} & \textbf{-3.96} & \multicolumn{1}{c|}{3.86$\times$} & \textbf{-3.81} & \multicolumn{1}{c|}{3.81$\times$} & \textbf{-2.04} & \multicolumn{1}{c|}{3.92$\times$} & \textbf{-3.16} & 4.19$\times$ \\ \midrule[1pt]
 & \textbf{Task} & \multicolumn{6}{c}{\textbf{POS Tagging}} & \multicolumn{4}{c}{\textbf{CWS}} \\ \midrule[1pt]
 & \multicolumn{1}{c|}{\multirow{2}{*}{\textbf{\begin{tabular}[c]{@{}c@{}}Dataset/\\ Model\end{tabular}}}} & \multicolumn{2}{c|}{\textbf{ARK Twitter}} & \multicolumn{2}{c|}{\textbf{CTB5 POS}} & \multicolumn{2}{c|}{\textbf{UD POS}} & \multicolumn{2}{c|}{\textbf{CTB5 Seg}} & \multicolumn{2}{c}{\textbf{UD Seg}} \\
 & \multicolumn{1}{c|}{} & Acc & \multicolumn{1}{c|}{Speedup} & F1($\Delta$) & \multicolumn{1}{c|}{Speedup} & F1($\Delta$) & \multicolumn{1}{c|}{Speedup} & F1($\Delta$) & \multicolumn{1}{c|}{Speedup} & F1($\Delta$) & Speedup \\ \midrule
 & \multicolumn{1}{c|}{BERT} & 91.51 & \multicolumn{1}{c|}{1.00$\times$} & 96.20 & \multicolumn{1}{c|}{1.00$\times$} & 95.04 & \multicolumn{1}{c|}{1.00$\times$} & 98.48 & \multicolumn{1}{c|}{1.00$\times$} & 98.04 & 1.00$\times$ \\
 & \multicolumn{1}{c|}{LSTM-CRF} & -0.92 & \multicolumn{1}{c|}{-} & -2.48 & \multicolumn{1}{c|}{-} & -6.30 & \multicolumn{1}{c|}{-} & -0.79 & \multicolumn{1}{c|}{-} & -3.03 & - \\ \midrule
\multicolumn{1}{c}{\parbox[t]{0mm}{\multirow{3}{*}{\rotatebox[origin=c]{90}{{$\sim$2X}}}}} & \multicolumn{1}{c|}{DistilBERT-6L} & -0.17 & \multicolumn{1}{c|}{2.00$\times$} & -0.27 & \multicolumn{1}{c|}{2.00$\times$} & -0.19 & \multicolumn{1}{c|}{2.00$\times$} & -0.12 & \multicolumn{1}{c|}{2.00$\times$} & -0.15 & 2.00$\times$ \\
 & \multicolumn{1}{c|}{SENTEE} & -0.49 & \multicolumn{1}{c|}{2.01$\times$} & -0.19 & \multicolumn{1}{c|}{1.95$\times$} & -0.35 & \multicolumn{1}{c|}{1.87$\times$} & \textbf{-0.02} & \multicolumn{1}{c|}{1.99$\times$} & -0.11 & 2.13$\times$ \\
 & \multicolumn{1}{c|}{TOKEE} & \textbf{-0.13} & \multicolumn{1}{c|}{1.96$\times$} & \textbf{-0.07} & \multicolumn{1}{c|}{2.04$\times$} & \textbf{-0.17} & \multicolumn{1}{c|}{1.93$\times$} & -0.05 & \multicolumn{1}{c|}{2.02$\times$} & \textbf{-0.07} & 2.10$\times$ \\ \midrule
\multicolumn{1}{c}{\parbox[t]{0mm}{\multirow{3}{*}{\rotatebox[origin=c]{90}{{$\sim$3X}}}}} & \multicolumn{1}{c|}{DistilBERT-4L} & -0.93 & \multicolumn{1}{c|}{3.00$\times$} & -0.74 & \multicolumn{1}{c|}{3.00$\times$} & -1.21 & \multicolumn{1}{c|}{3.00$\times$} & -0.19 & \multicolumn{1}{c|}{3.00$\times$} & -0.43 & 3.00$\times$ \\
 & \multicolumn{1}{c|}{SENTEE} & -1.98 & \multicolumn{1}{c|}{3.03$\times$} & -1.63 & \multicolumn{1}{c|}{3.12$\times$} & -2.47 & \multicolumn{1}{c|}{2.95$\times$} & -0.11 & \multicolumn{1}{c|}{3.03$\times$} & -0.47 & 2.99$\times$ \\
 & \multicolumn{1}{c|}{TOKEE} & \textbf{-0.67} & \multicolumn{1}{c|}{2.95$\times$} & \textbf{-0.30} & \multicolumn{1}{c|}{3.01$\times$} & \textbf{-1.04} & \multicolumn{1}{c|}{2.89$\times$} & \textbf{-0.06} & \multicolumn{1}{c|}{3.01$\times$} & \textbf{-0.20} & 3.01$\times$ \\ \midrule
\multicolumn{1}{c}{\parbox[t]{0mm}{\multirow{3}{*}{\rotatebox[origin=c]{90}{{$\sim$4X}}}}} & \multicolumn{1}{c|}{DistilBERT-3L} & \textbf{-1.04} & \multicolumn{1}{c|}{4.00$\times$} & -1.34 & \multicolumn{1}{c|}{4.00$\times$} & -4.16 & \multicolumn{1}{c|}{4.00$\times$} & -0.21 & \multicolumn{1}{c|}{4.00$\times$} & -1.01 & 4.00$\times$ \\
 & \multicolumn{1}{c|}{SENTEE} & -2.96 & \multicolumn{1}{c|}{3.98$\times$} & -2.37 & \multicolumn{1}{c|}{3.91$\times$} & -4.78 & \multicolumn{1}{c|}{3.96$\times$} & -0.83 & \multicolumn{1}{c|}{4.02$\times$} & -1.23 & 3.98$\times$ \\
 & \multicolumn{1}{c|}{TOKEE} & -2.17 & \multicolumn{1}{c|}{3.89$\times$} & \textbf{-1.29} & \multicolumn{1}{c|}{3.98$\times$} & \textbf{-4.03} & \multicolumn{1}{c|}{3.89$\times$} & \textbf{-0.14} & \multicolumn{1}{c|}{3.85$\times$} & \textbf{-0.53} & 3.97$\times$ \\ \bottomrule
\end{tabular}
    \caption{Comparison on BERT model. The performance of LSTM-CRF is from previous paper \citep{tian-etal-2020-joint-chinese,mengge-etal-2020-coarse1,TENER-hyan,gui-etal-2018-transferring}, and others are implemented by ourselves.}
    \label{main_exp_table}
  \end{table*}

\begin{table}[t]\scriptsize
\setlength{\tabcolsep}{2mm}
\begin{tabular}{@{}lccc@{}}
\toprule
\multicolumn{1}{c}{\textbf{Task}} & \textbf{NER} & \textbf{POS} & \textbf{CWS} \\ \midrule
\multicolumn{1}{c|}{\multirow{2}{*}{\textbf{\begin{tabular}[c]{@{}c@{}}Dataset/\\ Model\end{tabular}}}} & \multicolumn{1}{c|}{\textbf{Ontonotes4.0}} & \multicolumn{1}{c|}{\textbf{CTB5 POS}} & \textbf{UD Seg} \\
\multicolumn{1}{c|}{} & \multicolumn{1}{c|}{F1($\Delta$)/Speedup} & \multicolumn{1}{c|}{F1($\Delta$)/Speedup} & F1($\Delta$)/Speedup \\ \midrule
\multicolumn{4}{c}{\textbf{RoBERTa-base}} \\ \midrule
\multicolumn{1}{l|}{RoBERTa} & \multicolumn{1}{c|}{79.32 / 1.00$\times$} & \multicolumn{1}{c|}{96.29 / 1.00$\times$} & 97.91 / 1.00$\times$ \\ \midrule
\multicolumn{1}{l|}{RoBERTa 6L} & \multicolumn{1}{c|}{-4.03 / 2.00$\times$} & \multicolumn{1}{c|}{-0.11 / 2.00$\times$} & -0.28 / 2.00$\times$ \\
\multicolumn{1}{l|}{SENTEE} & \multicolumn{1}{c|}{-0.55 / 1.98$\times$} & \multicolumn{1}{c|}{-0.08 / 1.85$\times$} &  -0.08 / 2.01$\times$\\
\multicolumn{1}{l|}{TOKEE} & \multicolumn{1}{c|}{\textbf{-0.19} / 2.01$\times$} & \multicolumn{1}{c|}{\textbf{-0.06} / 1.98$\times$} & \textbf{-0.02} / 1.97$\times$\\ \midrule
\multicolumn{1}{l|}{RoBERTa 4L} & \multicolumn{1}{c|}{-11.97 / 3.00$\times$} & \multicolumn{1}{c|}{-1.59 / 3.00$\times$} & -0.76 / 3.00$\times$ \\
\multicolumn{1}{l|}{SENTEE} & \multicolumn{1}{c|}{-5.61 / 3.03$\times$} & \multicolumn{1}{c|}{-1.02 / 2.80$\times$} &  -0.51 / 2.99$\times$\\
\multicolumn{1}{l|}{TOKEE} & \multicolumn{1}{c|}{\textbf{-0.58} / 3.05$\times$} & \multicolumn{1}{c|}{\textbf{-0.51} / 3.01$\times$} & \textbf{-0.12} / 3.00$\times$ \\ \midrule
\multicolumn{4}{c}{\textbf{ALBERT-base}} \\ \midrule
\multicolumn{1}{l|}{ALBERT} & \multicolumn{1}{c|}{76.24 / 1.00$\times$} & \multicolumn{1}{c|}{95.73 / 1.00$\times$} & 97.00 / 1.00$\times$ \\ \midrule
\multicolumn{1}{l|}{ALBERT 6L} & \multicolumn{1}{c|}{-5.78 / 2.00$\times$} & \multicolumn{1}{c|}{-0.42 / 2.00$\times$} & -0.20 / 2.00$\times$ \\
\multicolumn{1}{l|}{SENTEE} & \multicolumn{1}{c|}{-1.25 / 2.01$\times$} & \multicolumn{1}{c|}{-0.23 / 1.97$\times$} &   \textbf{-0.02} / $1.98\times$ \\
\multicolumn{1}{l|}{TOKEE} & \multicolumn{1}{c|}{\textbf{-0.43} / 1.97$\times$} & \multicolumn{1}{c|}{\textbf{-0.08} / 2.00$\times$} & -0.03 / 2.01$\times$ \\ \midrule
\multicolumn{1}{l|}{ALBERT 4L} & \multicolumn{1}{c|}{-12.93 / 3.00$\times$} & \multicolumn{1}{c|}{-2.15 / 3.00$\times$} & -1.38 / 3.00$\times$ \\
\multicolumn{1}{l|}{SENTEE} & \multicolumn{1}{c|}{-6.90 / 2.83$\times$} & \multicolumn{1}{c|}{-1.47 / 3.11$\times$} &   -0.41 / 2.97$\times$ \\
\multicolumn{1}{l|}{TOKEE} & \multicolumn{1}{c|}{\textbf{-0.87} / 2.87$\times$} & \multicolumn{1}{c|}{\textbf{-0.58} / 2.89$\times$} & \textbf{-0.06} / 2.98$\times$ \\ \bottomrule
\end{tabular}
\caption{Result on RoBERTa and ALBERT.}
\label{main_exp_table_2}
\end{table}

\section{Experiment}

\subsection{Computational Cost Measure}
We use average floating-point operations (FLOPs) as the measure of computational cost, which denotes how many floating-point operations the model performs for a single sample. The FLOPs is universal enough since it is not involved with the model running environment (CPU, GPU or TPU) and it can measure the theoretical running time of the model. In general, the lower the model's FLOPs is, the faster the model's inference is.
\subsection{Experimental Setup}
\subsubsection{Dataset}
To verify the effectiveness of our methods, We conduct experiments on ten English and Chinese datasets of sequence labeling, covering \textbf{NER}: CoNLL2003 \citep{conll-dataset}, Twitter NER \citep{twitter-ner-dataset}, Ontonotes 4.0 (Chinese) \citep{ontonotes4-dataset}, Weibo \citep{weibo1,weibo2} and CLUE NER \citep{CLUENER}, \textbf{POS}: ARK Twitter \citep{ark1,ark2}, CTB5 POS \citep{ctb5} and UD POS \citep{ud-dataset}, \textbf{CWS}: CTB5 Seg \citep{ctb5} and UD Seg \citep{ud-dataset}. Besides the standard benchmark dataset like CoNLL2003 and Ontonotes 4.0, we also choose some datasets closer to real-world application to verify the actual utility of our methods, such as Twitter NER and Weibo in social media domain. We use the same dataset preprocessing and split as in previous work \citep{LSTM-CRF-1,mengge-etal-2020-coarse1,jia-etal-2020-entity,tian-etal-2020-joint-chinese,nguyen-etal-2020-bertweet}.

\subsubsection{Baseline}
We compare our methods with three baselines:
\begin{itemize}
\setlength{\itemsep}{1pt}%
\setlength{\parskip}{1pt}%
    \item \textbf{BiLSTM-CRF} \citep{LSTM-CRF-1,ma-hovy-2016-end} The most widely used model in sequence labeling tasks before the pre-trained language model prevails in NLP.
    \item \textbf{BERT} The powerful stacked Transformer encoder model, pre-trained on large-scale corpus, which we use as the backbone of our methods.
    \item \textbf{DistilBERT} The most well-known distillation method of BERT. Huggingface released 6 layers DistilBERT for English \citep{sanh2019distilbert}. For comparison, we distill \{3, 4\} and \{3, 4, 6\} layers DistilBERT for English and Chinese using the same method.
\end{itemize}
\subsubsection{Hyper-Parameters}
For all datasets, We use batch size=10. We perform grid search over learning rate in \{5e-6,1e-5,2e-5\}. We choose learning rate and the model based on the development set. We use the AdamW optimizer \citep{adamw}. The warmup step, weight decay is set to 0.05, 0.01, respectively.

\subsection{Main Results}
For English Datasets, we use the `BERT-base-cased' released by Google \citep{devlin-etal-2019-bert} as backbone. For Chinese Datasets, we use `BERT-wwm' released by \citep{chinese-bert-wwm}. The DistilBERT is distilled from the backbone BERT.

To fairly compare our methods with baselines, we turn the speedup ratio of our methods to be consistent with the corresponding static baseline. We report the average performance over 5 times under different random seeds. The overall results are shown in Table \ref{main_exp_table}, where the \textit{speedup} is based on the backbone. We can see both SENTEE and TOKEE brings little performance drop and outperforms DistilBERT in speedup ratio of 2, which has achieved similar effect like existing early-exit for text classification. Under higher speedup, 3$\times$ and 4$\times$, SENTEE shows its weakness but TOKEE can still keep a certain performance. And under 2$\sim$4$\times$ speedup ratio, TOKEE has a lower performance drop than DistilBERT. What's more, for datasets where BERT can show its power than LSTM-CRF, e.g., Chinese NER, TOKEE (4$\times$) on BERT can still outperform LSTM-CRF significantly. This indicates the potential utility of it in complicated real-world scenario.

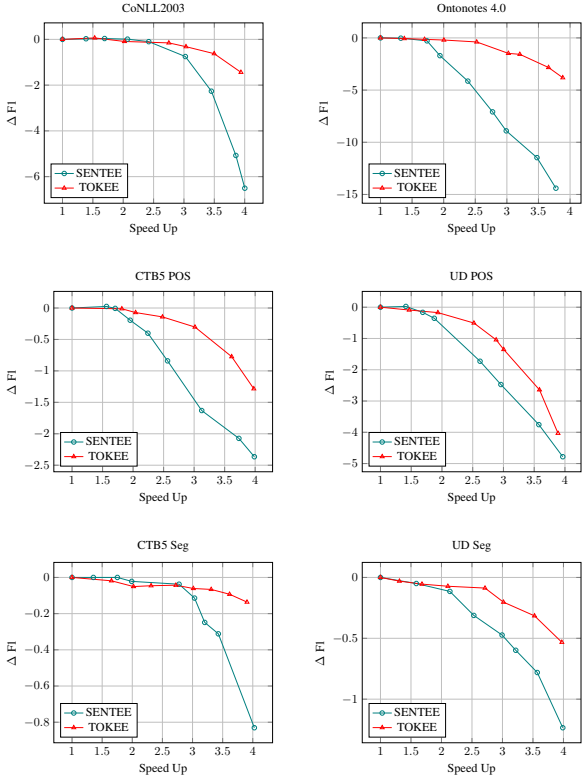
\begin{figure}[t!]
  \centering
    \subfigure{
  \centering
  \begin{tikzpicture}[scale=0.42]
    \begin{axis}[
      xlabel=Speed Up,
      ylabel=$\Delta$ F1,
      grid=major,
      title=CoNLL2003,
       legend pos=south west
    ]

    \addplot [color=teal,mark=o]coordinates { %
          (1, 0)
    (1.387223,0.0276)
    (1.693254,0.0372)
    (2.070523,0.0101)
    (2.42059,-0.10)
    (3.02312, -0.75)
    (3.452127,-2.2654)
    (3.855173,-5.07448)
    (4.002062,-6.5059)
    };
    \addlegendentry{SENTEE}

    \addplot [color=red,mark=triangle]coordinates { %
      (1, 0)
      (1.530423,0.0639)
      (2.020456,-0.0914)
      (2.751263,-0.1538)
        (3.033428,-0.3195)
        (3.497519,-0.6248)
      (3.93826,-1.4393)
    };
    \addlegendentry{TOKEE}

    \end{axis}
  \end{tikzpicture}}\hfill
    \subfigure{
  \centering
  \begin{tikzpicture}[scale=0.42]
    \begin{axis}[
      xlabel=Speed Up,
      ylabel=$\Delta$ F1,
      grid=major,
      title=Ontonotes 4.0,
       legend pos=south west
    ]

    \addplot [color=teal,mark=o]coordinates { %
         (1, 0)
        (1.321817,-0.0252)
        (1.735822,-0.2822)
        (1.940977,-1.7004)
        (2.382629,-4.14)
        (2.773446,-7.0855)
        (2.990167,-8.905)
         (3.476334, -11.472)
        (3.775921,-14.4003)
    };
    \addlegendentry{SENTEE}

    \addplot [color=red,mark=triangle]coordinates { %
        (1, 0)
        (1.377202,-0.0629)
        (1.697135,-0.1385)
      (2.003233,-0.2034)
      (2.522671,-0.3884)
      (3.023897,-1.4826)
      (3.204999,-1.5729)
      (3.659528,-2.8356)
      (3.884669,-3.8147)
    };
    \addlegendentry{TOKEE}

    \end{axis}
  \end{tikzpicture}}\hfill
      \subfigure{
  \centering
  \begin{tikzpicture}[scale=0.42]
    \begin{axis}[
      xlabel=Speed Up,
      ylabel=$\Delta$ F1,
      grid=major,
      title=CTB5 POS,
       legend pos=south west
    ]

    \addplot [color=teal,mark=o]coordinates { %
      (1, 0)
      (1.563894,0.0249)
      (1.706599,-0.006)
      (1.952734,-0.1943)
      (2.243169,-0.4009)
      (2.56455,-0.8405)
      (3.12455,-1.6305)
      (3.733424, -2.0733)
      (3.986817,-2.3652)
    };
    \addlegendentry{SENTEE}

    \addplot [color=red,mark=triangle]coordinates { %
      (1, 0)
      (1.816122,-0.0124)
      (2.041353,-0.0725)
      (2.481885,-0.1415)
      (3.013776,-0.3029)
      (3.615447,-0.774)
      (3.977919,-1.2845)
    };
    \addlegendentry{TOKEE}

    \end{axis}
  \end{tikzpicture}}\hfill
     \subfigure{
  \centering
  \begin{tikzpicture}[scale=0.42]
    \begin{axis}[
      xlabel=Speed Up,
      ylabel=$\Delta$ F1,
      grid=major,
      title=UD POS,
       legend pos=south west
    ]

    \addplot [color=teal,mark=o]coordinates { %
      (1, 0)
      (1.413139,0.0246)
      (1.687621,-0.1642)
      (1.875007,-0.3542)
      (2.619259, -1.7283)
      (2.958467,-2.4726)
      (3.575478,-3.755)
      (3.964888,-4.782)
    };
    \addlegendentry{SENTEE}

    \addplot [color=red,mark=triangle]coordinates { %
        (1, 0)
        (1.467743,-0.09)
        (1.933993,-0.1727)
        (2.519487,-0.506)
        (2.881283,-1.0439)
      (3.004409,-1.3524)
      (3.585969,-2.6425)
      (3.887334,-4.0319)
    };
    \addlegendentry{TOKEE}

    \end{axis}
  \end{tikzpicture}}\hfill
        \subfigure{
  \centering
  \begin{tikzpicture}[scale=0.42]
    \begin{axis}[
      xlabel=Speed Up,
      ylabel=$\Delta$ F1,
      grid=major,
      title=CTB5 Seg,
       legend pos=south west
    ]

    \addplot [color=teal,mark=o]coordinates { %
            (1, 0)
      (1.353872,0)
      (1.75235,0)
      (1.99,-0.0218)
      (2.776266,-0.0374)
      (3.033459,-0.1141)
            (3.202468, -0.2486)
      (3.428327,-0.3115)
      (4.026164,-0.8305)
    };
    \addlegendentry{SENTEE}

    \addplot [color=red,mark=triangle]coordinates { %
      (1, 0)
      (1.654165,-0.0186)
      (2.0204,-0.0504)
      (2.311866,-0.0458)
      (2.723395,-0.0429)
      (3.013877,-0.0611)
      (3.309531,-0.0662)
      (3.614824,-0.0929)
      (3.904, -0.1362)
    };
    \addlegendentry{TOKEE}

    \end{axis}
  \end{tikzpicture}}\hfill
     \subfigure{
  \centering
  \begin{tikzpicture}[scale=0.42]
    \begin{axis}[
      xlabel=Speed Up,
      ylabel=$\Delta$ F1,
      grid=major,
      title=UD Seg,
       legend pos=south west
    ]

    \addplot [color=teal,mark=o]coordinates { %
        (1, 0)
        (1.586765,-0.0498)
      (2.137886,-0.1148)
      (2.530306,-0.312)
      (2.993365,-0.4731)
        (3.217802, -0.5985)
      (3.570536,-0.7811)
      (3.98886,-1.2348)
    };
    \addlegendentry{SENTEE}

    \addplot [color=red,mark=triangle]coordinates { %
        (1, 0)
        (1.310479,-0.0302)
        (1.680387,-0.0548)
        (2.104964,-0.0724)
        (2.713778,-0.0873)
        (3.017166,-0.2031)
        (3.525794,-0.3156)
        (3.973308,-0.5325)
    };
    \addlegendentry{TOKEE}

    \end{axis}
  \end{tikzpicture}}\hfill

  \caption{The performance-speedup trade-off curve on six datasets of SENTEE and TOKEE.}
  \label{performance-speedup-trade-off-fig}

\end{figure}

To explore the fine-grained performance change under different speedup ratio, We visualize the speedup-performance trade-off curve on 6 datasets, in Figure\ref{performance-speedup-trade-off-fig}. We observe that,
\begin{itemize}
\setlength{\itemsep}{1pt}%
\setlength{\parskip}{1pt}%
    \item Before the speedup ratio rises to a certain turning point, there is almost no drop on performance. After that, the performance will drop gradually. This shows our methods keep the superiority of existing early-exit methods \citep{xin-etal-2020-deebert}.
    \item As the speedup rises, TOKEE will encounter the speedup turning point later than SENTEE. After both methods reach the turning point, SENTEE's performance degradation is more drastic than TOKEE. These both indicate the higher speedup ceiling of TOKEE.
    \item On some datasets, such as CoNLL2003, we observe a little performance improvement under low speedup ratio, we attribute this to the potential regularization brought by early-exit, such as alleviating overthinking \citep{shallow-deep-overthinking}.
\end{itemize}

To verify the versatility of our method over different PTMs, we also conduct experiments on two well-known BERT variants, RoBERTa \citep{liu2019roberta}\footnote{\href{https://github.com/ymcui/Chinese-BERT-wwm}{https://github.com/ymcui/Chinese-BERT-wwm}.} and ALBERT \citep{Lan2020ALBERT}\footnote{\href{https://github.com/brightmart/albert\_zh}{https://github.com/brightmart/albert\_zh}.}, as shown in Table \ref{main_exp_table_2}. We can see that SENTEE and TOKEE also significantly outperform static backbone internal layer on three Representative datasets of corresponding tasks. For RoBERTa and ALBERT, we also observe the TOKEE can have a better performance than SENTEE under high speedup ratio.

 \begin{figure}
 \centering
     \subfigure{
  \centering
  \begin{tikzpicture}[scale=0.75]
  \centering
    \begin{axis}[
    yscale=0.8,
      xlabel=$k$,
      ylabel=$\Delta$ F1,
      grid=major,
       legend style={at={(0.95,0.7)}},
    ]

    \addplot [color=teal,mark=o]coordinates { %
        (0,-0.7887)
        (1,-0.07)
        (2,-0.1677)
        (3,-0.0795)
        (4,-0.0052)
        (5,-0.0081)
        (6,0.0191)
        (7,-0.095)
        (8,0.0721)
        (9,-0.0435)
        (10,-0.0435)
        (11,-0.053)
        (12,-0.0611)
    };
    \addlegendentry{Speedup $\sim 2\times$}

    \addplot [color=red,mark=triangle]coordinates { %
        (0,-2.7722)
        (1,-0.6186)
        (2,-0.482)
        (3,-0.508)
        (4,-0.7807)
        (5,-1.0001)
        (6,-1.152)
        (7,-1.3355)
        (8,-1.4121)
        (9,-1.3066)
        (10,-1.2714)
        (11,-1.3146)
        (12,-1.8533)
    };
    \addlegendentry{Speedup $\sim 3\times$}

    \addplot [color=orange,mark=square]coordinates { %
        (0,-6.6919)
        (1,-3.7256)
        (2,-3.0897)
        (3,-4.4393)
        (4,-4.9471)
        (5,-6.0191)
        (6,-6.5232)
        (7,-6.7722)
        (8,-7.1347)
        (9,-7.2985)
        (10,-7.1711)
        (11,-7.7283)
        (12,-7.711)
    };
    \addlegendentry{Speedup $\sim 4\times$}

    \end{axis}
  \end{tikzpicture}}\hfill
  \caption{The performance change over different window size under the same speedup ratio.}
  \label{effect-of-window-fig}
 \end{figure}
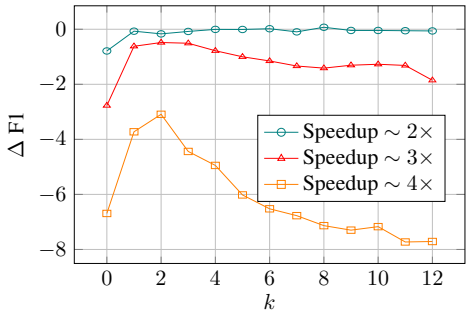

\subsection{Analysis}
In this section, we conduct a set of detailed analysis on our methods.

\subsubsection{The Effect of Window Size}
We show the performance change under different $k$ in Figure \ref{effect-of-window-fig}, keeping the speedup ratio consistent. We observe that: (1) when $k$ is 0, in other words, not using window-based uncertainty but token-independent uncertainty, the performance is the almost lowest across different speedup ratio, because it does not consider local dependency at all. This shows the necessity of the window-based uncertainty. (2) When $k$ is relatively large, it will bring significant performance drop under high speedup ratio (3$\times$ and 4$\times$), like SENTEE. (3) It is necessary to choose an appropriate $k$ under high speedup ratio, where the effect of different $k$ has a high variance.

\begin{figure}
    \centering
    \scalefont{0.5}
    \centering

    \begin{tikzpicture}[scale=0.7]
  \begin{axis}[
    ybar,
    scale only axis,
    yscale=0.3,
    xscale=1.1,
    enlargelimits=0.15,
    legend style={at={(0.5,-0.2)},
      anchor=north,legend columns=-1},
    ylabel={Off-Ramp 8 Acc},
    symbolic x coords={0$\sim$0.1,0.1$\sim$0.2,0.2$\sim$0.3,%
		0.3$\sim$0.4,0.4$\sim$0.5,0.5$\sim$0.6,0.6$\sim$0.7,0.7$\sim$0.8,0.8$\sim$0.9,0.9$\sim$1},
    xtick=data,
    nodes near coords,
	nodes near coords align={vertical},
    x tick label style={rotate=30,anchor=east},
    ]
    y tick label style={
        /pgf/number format/.cd,
        fixed,
        fixed zerofill,
        precision=0,
        /tikz/.cd
    }
    \addplot[fill=royalblue] coordinates {(0$\sim$0.1,96.72) (0.1$\sim$0.2,70.99)
		(0.2$\sim$0.3,70.19) (0.3$\sim$0.4,62.72) (0.4$\sim$0.5,50)(0.5$\sim$0.6,46.15) (0.6$\sim$0.7,40)
		(0.7$\sim$0.8,30.59) (0.8$\sim$0.9,26.3) (0.9$\sim$1,22.3)};
  \end{axis}
\end{tikzpicture}

    \subfigure[Uncertainty]{
    \centering

    \begin{tikzpicture}[scale=0.7]
  \begin{axis}[
    ybar,
    scale only axis,
    yscale=0.3,
    xscale=1.1,
    enlargelimits=0.15,
    ylabel={Off-Ramp 4 Acc},
    symbolic x coords={0$\sim$0.1,0.1$\sim$0.2,0.2$\sim$0.3,%
		0.3$\sim$0.4,0.4$\sim$0.5,0.5$\sim$0.6,0.6$\sim$0.7,0.7$\sim$0.8,0.8$\sim$0.9,0.9$\sim$1},
    xtick=data,
    nodes near coords,
	nodes near coords align={vertical},
    x tick label style={rotate=30,anchor=east},
    y tick label style={
        /pgf/number format/.cd,
        fixed,
        fixed zerofill,
        precision=0,
        /tikz/.cd
    }
    ]
    \addplot[fill=royalblue] coordinates {(0$\sim$0.1,98.43) (0.1$\sim$0.2,97.79)
		(0.2$\sim$0.3,93.54) (0.3$\sim$0.4,86.69) (0.4$\sim$0.5,80.8)(0.5$\sim$0.6,70.15) (0.6$\sim$0.7,61.51)
		(0.7$\sim$0.8,57.49) (0.8$\sim$0.9,52.66) (0.9$\sim$1,36.84)};
  \end{axis}
\end{tikzpicture}
\label{uncertainty-acc-fig}

    }

\begin{tikzpicture}[scale=0.7]
\begin{axis}[
    ybar,
    scale only axis,
    yscale=0.3,
    xscale=1.1,
    xtick=data,
	ylabel={Off-Ramp 8 Acc},
	enlargelimits=0.15,
	ybar=5pt,%
	bar width=9pt,
	nodes near coords,
	    y tick label style={
        /pgf/number format/.cd,
        fixed,
        fixed zerofill,
        precision=0,
        /tikz/.cd
    }
]
\addplot[fill=royalblue]
	coordinates {(0,93.83) (1,94.43)
		 (2,94.56) (3,94.64) (4,94.7)
		 (5,94.78) (6,94.79) (7,94.8) (8,94.8) (9,94.81)};

\end{axis}
\end{tikzpicture}

\subfigure[$k$]{
\begin{tikzpicture}[scale=0.7]
\begin{axis}[
    ybar,
    scale only axis,
    yscale=0.3,
    xscale=1.1,
	ylabel={Off-Ramp 4 Acc},
	enlargelimits=0.15,
	xtick=data,
	ybar=5pt,%
	bar width=9pt,
	nodes near coords,
		    y tick label style={
        /pgf/number format/.cd,
        fixed,
        fixed zerofill,
        precision=0,
        /tikz/.cd
    }
]
\addplot[fill=royalblue]
	coordinates {(0,94.69) (1,96.72)
		 (2,96.96) (3,97.15) (4,97.28)
		 (5,97.42) (6,97.6) (7,97.64) (8,97.63) (9,97.67)};

\end{axis}
\end{tikzpicture}
\label{k-acc-fig}
}

    \caption{The relation of off-ramps' accuracy between uncertainty and window size. Two off-ramps of shallow and deep layers are analyzed on CoNLL2003.}
    \label{acc-fig}
\end{figure}
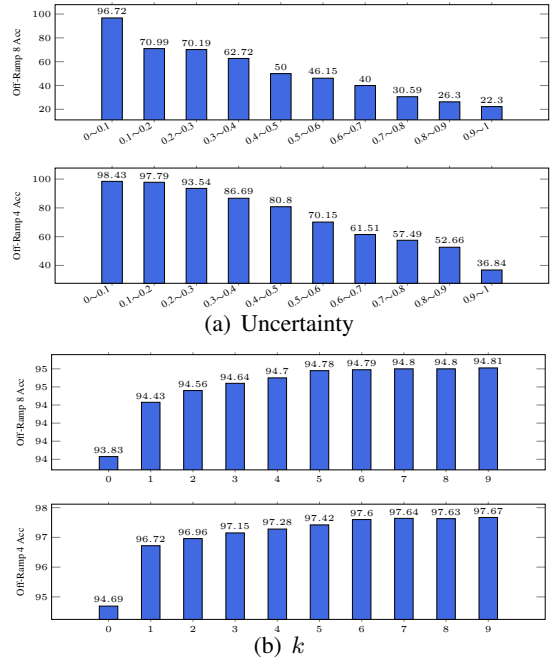

\subsubsection{Accuracy V.S. Uncertainty} \citet{liu-etal-2020-fastbert} verified `the lower the uncertainty, the higher the accuracy' on text classification. Here, we'd like to verify our window-based uncertainty on sequence labeling. In detail, we verify the entire window-based uncertainty and its specific hyper-parameter, $k$, on CoNLL2003, shown in Figure \ref{acc-fig}. For the uncertainty, we intercept the 4 th and 8 th off-ramps and calculate their accuracy in each uncertainty interval, when $k$=2. The result shown in Figure \ref{uncertainty-acc-fig} indicates that `the lower the window-based uncertainty, the higher the accuracy', similar as in text classification. For $k$, we set a certain threshold = 0.3, and calculate accuracy of tokens whose window-based uncertainty is small than the threshold under different $k$, shown in Figure \ref{k-acc-fig}. The result shows that, as $k$ increases: (1) The accuracy of screened tokens is higher. This shows that the wider of a token's low-uncertainty neighborhood, the more accurate the token's prediction is. This also verifies the validity of window-based uncertainty strategy. (2) The accuracy improvement slows down. This shows the low relevance of distant tokens' uncertainty and explains why large $k$ performs not well under high speedup ratio: it does not help improving more accurate exiting but slowing down exiting.

\begin{figure}[t]
    \centering
    \scalefont{0.7}
\begin{tikzpicture} []
\begin{axis}[
    scale only axis,
    yscale=0.4,
    xscale=0.8,
	xlabel={Sentence Length},
	ylabel={Speedup Ratio},
	xtick={-0.3,0.7,1.7,2.7,3.7},
	xticklabels={10,30,50,70,$\textgreater$ 70},
    ymin=1,
    ymax=4,
    legend cell align={left},
	grid=major,
	legend pos=south west,
	width=\linewidth,
    height=0.8\linewidth,
]

\addplot+[color=red,mark=square] coordinates{

(-0.3, 2.949832)
(0.7, 3.102688)
(1.7, 3.102688)
(2.7, 2.918125)
(3.7, 3.163033)
};
\addlegendentry{$k=2$}

\addplot+[color=teal,mark=star] coordinates{
(-0.3, 2.772411)
(0.7, 2.475589)
(1.7, 2.295665)
(2.7, 1.843588)
(3.7, 1.651127)
};
\addlegendentry{$k=+\infty$}

\end{axis}
\end{tikzpicture}
    \caption{Influence of sentence length.}
    \label{effect-sentence-length-fig}
\end{figure}
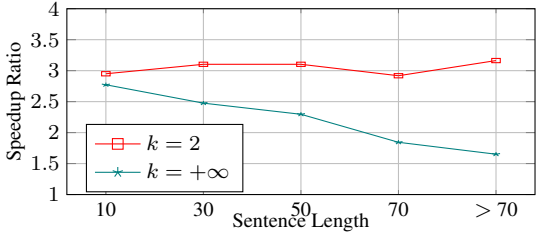

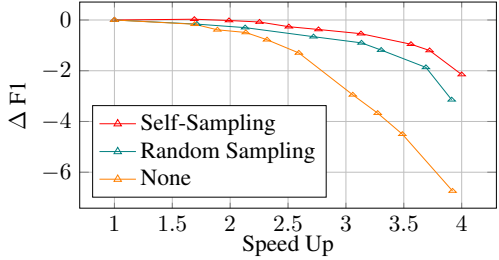
\begin{figure}
    \centering
        \subfigure{
  \centering
  \begin{tikzpicture}[scale=0.8]
    \begin{axis}[
    legend cell align={left},
        yscale=0.6,
      xlabel=Speed Up,
      ylabel=$\Delta$ F1,
      grid=major,
       legend pos=south west
    ]

    \addplot [color=red,mark=triangle]coordinates { %
      (1, 0)
      (1.690456,0.03)
      (1.99, -0.02)
        (2.253428,-0.0795)
      (2.506826,-0.264)
      (2.763013,-0.3743)
      (3.13, -0.54)
      (3.56181,-0.9555)
      (3.722966,-1.2019)
      (4.0, -2.15)
    };
    \addlegendentry{Self-Sampling}

        \addplot [color=teal,mark=triangle]coordinates { %
      (1, 0)
      (1.712373,-0.1627)
      (2.129454,-0.301)
      (2.720146,-0.6583)
      (3.134402,-0.9017)
      (3.304386,-1.1856)
      (3.691171,-1.8689)
      (3.913778,-3.1448)
    };
    \addlegendentry{Random Sampling}

    \addplot [color=orange,mark=triangle]coordinates { %
        (1,0)
        (1.692447,-0.1684)
        (1.88966,-0.3869)
        (2.134494,-0.4925)
        (2.31841,-0.7787)
        (2.593317,-1.2994)
        (3.059475,-2.9504)
        (3.274394,-3.6722)
        (3.487893,-4.5023)
        (3.921523,-6.7403)

    };
    \addlegendentry{None}

    \end{axis}
  \end{tikzpicture}}\hfill
    \caption{Comparison between different training way.}
    \label{cmp-self-sample-train-fig}
\end{figure}

\subsubsection{Influence of Sequence Length}
Transformer-based PTMs, e.g. BERT, face a challenge in processing long text, due to the $O(N^2d)$ computational complexity brought by self-attention. Since the TOKEE reduces the layer-wise computational complexity from $O(N^2d+Nd^2)$ to $O(NMd+Md^2)$ and SENTEE does not, we'd like to explore their effect over different sentence length. We compare the highest speedup ratio of TOKEE and SENTEE when performance drop $< 1$ on Ontonotes 4.0, shown in Figure \ref{effect-sentence-length-fig}. We observe that TOKEE has a stable computational cost saving as the sentence length increases, but SENTEE's speedup ratio will gradually reduce. For this, we give an intuitive explanation. In general, a longer sentence has more tokens, it is more difficult for the model to give them all confident prediction at the same layer. This comparison reveals the potential of TOKEE on accelerating long text inference.

\begin{figure}[t]
    \centering
    \subfigure{    \includegraphics[width=0.45\textwidth]{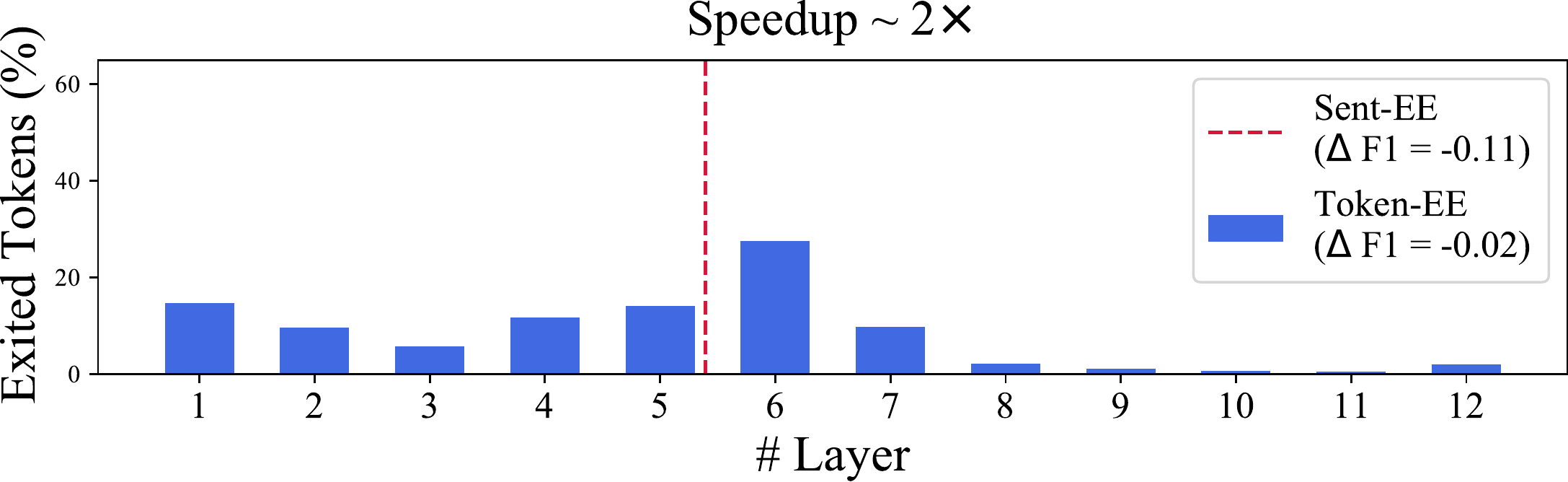}}
    \subfigure{    \includegraphics[width=0.45\textwidth]{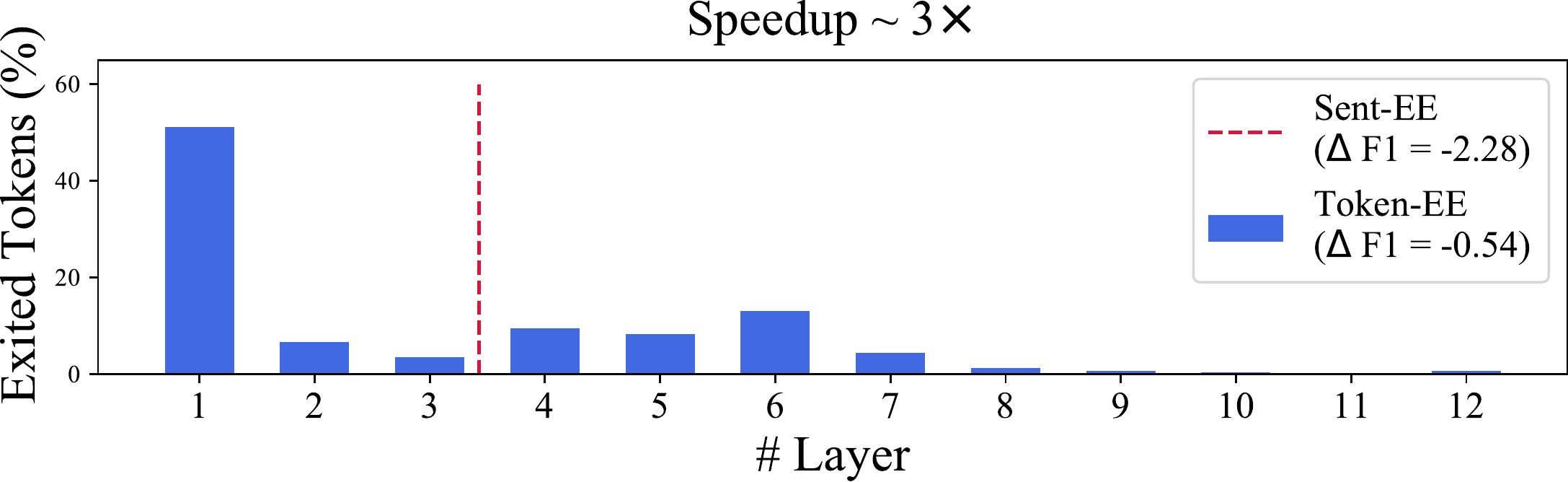}}
    \subfigure{    \includegraphics[width=0.45\textwidth]{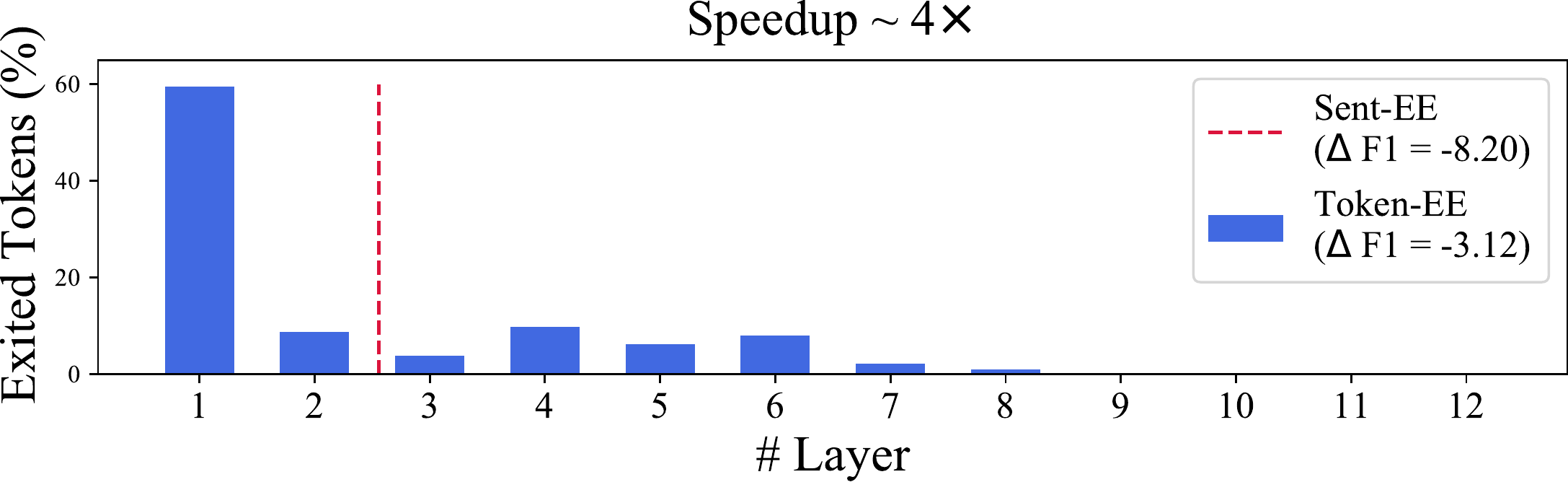}}
    \caption{TOKEE layer distribution in a sentence.}
    \label{layer-distribution}
\end{figure}

\subsubsection{Effects of Self-Sampling Fine-Tuning}
\label{self-sampling-section}
To verify the effect of self-sampling fine-tuning in Section \ref{sec:ft-tokenEE}, we compare it with random sampling and no extra fine-tuning on CoNLL2003. The performance-speedup trade-off curve of TOKEE is shown in Figure \ref{cmp-self-sample-train-fig}, which shows self-sampling  is always better than random sampling for TOKEE. As speedup ratio rises, this trend is more significant. This shows the self-sampling can help more in reducing the gap of training and inference. As for no extra fine-tuning, it will deteriorate drastically at high speedup ratio. But it can roughly keep a certain capability at low speedup ratio, which we attribute to the residual-connection of PTM and similar results were reported by \citet{residual-ensemble}.
\subsubsection{Layer Distribution of Early-Exit}
In TOKEE, by halt-and-copy mechanism, each token goes through a different number of PTM layers according to the difficulty. We show the average distribution of a sentence's tokens exiting layers under different speedup ratio on CoNLL2003, in Figure \ref{layer-distribution}. We also draw the average exiting layer number of SENTEE under the same speedup ratio. We observe that as speedup ratio rises, more tokens will exit at the earlier layer but a bit of tokens can still go through the deeper layer even when 4$\times$, meanwhile, the SENTEE's average exiting layer number reduces to ~2.5, where the PTM's encoding power is severely cut down. This gives an intuitive explanation of why TOKEE is more effective than SENTEE under high speedup ratio: although both SENTEE and TOKEE can dynamically adjust computational cost on the sample-level, TOKEE can adjust do it in a more fine-grained way.

\section{Related Work}
PTMs are powerful but have high computational cost. To accelerate them, many attempts have been made. A kind of methods is to reduce its size, such as distillation \cite{sanh2019distilbert,tiny-bert}, structural pruning \cite{prune-head,layer-drop} and quantization \cite{Shen_Dong_Ye_Ma_Yao_Gholami_Mahoney_Keutzer_2020}.

Another kind of methods is early-exit, which dynamically adjusts the encoding layer number of different samples ~\citep{liu-etal-2020-fastbert,xin-etal-2020-deebert,schwartz-etal-2020-right,zhou-bert-patience,DBLP:journals/corr/abs-2012-14682}.
While they introduced early-exit mechanism in simple classification tasks, our methods are proposed for the more complicated scenario: sequence labeling, where it has not only one prediction probability and it's necessary to consider the dependency of token exitings. \citet{jiatao-dpeth-adaptive} proposed Depth-Adaptive Transformer to accelerate machine translation. However, their early-exit mechanism is designed for auto-regressive sequence generation, in which the exit of tokens must be in left-to-right order. Therefore, it is unsuitable for language understanding tasks. %
Different from their method, our early-exit mechanism can consider the exit of all tokens simultaneously.

\section{Conclusion and Future Work}
In this work, we propose two early-exit mechanisms for sequence labeling: SENTEE and TOKEE. The former is a simple extension of sequence-level early-exit while the latter is specially designed for sequence labeling, which can conduct more fine-grained computational cost allocation. We equip TOKEE with window-based uncertainty and self-sampling finetuning to make it more robust and faster. The detailed analysis verifies their effectiveness. SENTEE and TOKEE can achieve 2$\times$ and 3$\sim$4$\times$ speedup with minimal performance drop.

For future work, we wish to explore: (1) leveraging the exited token's label information to help the exiting of remained tokens; (2) introducing CRF or other global decoding methods into early-exit for sequence labeling.

\section*{Acknowledgments}
We thank anonymous reviewers for their detailed reviews and great suggestions. This work was supported by the National Key Research and Development Program of China (No. 2020AAA0106700), National Natural Science Foundation of China (No. 62022027) and Major Scientific Research Project of Zhejiang Lab (No. 2019KD0AD01).

\bibliographystyle{acl_natbib}
\bibliography{myref}

  \end{CJK}
\end{document}